\documentclass[letterpaper, 10 pt, journal, twoside]{IEEEtran}
\usepackage{graphics} % for pdf, bitmapped graphics files
\usepackage{graphicx}
\usepackage{epsfig} % for postscript graphics files
\usepackage{times} % assumes new font selection scheme installed
\usepackage{amsmath} % assumes amsmath package installed
\usepackage{amssymb}  % assumes amsmath package installed
\usepackage{rotating}
\usepackage{array}
\usepackage{multirow}
\usepackage{makecell}
\usepackage{xcolor}

\usepackage{multicol}
\usepackage{booktabs}
\usepackage{makecell}
\usepackage{colortbl}
\usepackage{hyperref}
\usepackage{algorithm}
\usepackage{algorithmic}
\usepackage{marvosym}

\usepackage{geometry}
\begin{document}
%
% Do not put math or special symbols in the title.
% \title{What is the Best Feature for Multi-View Fine-grained Bimanual Manipulation}

% \title{Which View is the Best? Dynamic View Transfer for Multi-View Fine-grained Manipulation}

\title{\LARGE \bf
BFA: Best-Feature-Aware Fusion for Multi-View \\Fine-grained Manipulation
}

\author{Zihan Lan$^{1*}$ Weixin Mao$^{2*\dagger}$ Haosheng Li$^{3*}$ 
Le Wang$^{4}$  Tiancai Wang$^{1}$  Haoqiang Fan$^{1}$ Osamu Yoshie$^{2\ddagger}$ % <-this % stops a space
% \thanks{*This work was not supported by any organization}% <-this % stops a space
% \thanks{$^{1}$Albert Author is with Faculty of Electrical Engineering, Mathematics and Computer Science,
%         University of Twente, 7500 AE Enschede, The Netherlands
%         {\tt\small albert.author@papercept.net}}%
% \thanks{$^{2}$Bernard D. Researcheris with the Department of Electrical Engineering, Wright State University,
%         Dayton, OH 45435, USA
%         {\tt\small b.d.researcher@ieee.org}}%
% }
\thanks{$^{1}$ MEGVII Technology, Beijing, China
        % {\tt\small albert.author@papercept.net}
        }
\thanks{$^{2}$ Waseda University, Tokyo, Japan
        % {\tt\small albert.author@papercept.net}
        }
\thanks{$^{3}$ Institute of Software, Chinese Academy of Sciences, Beijing, China
         % {\tt\small (lihaosheng23,)@mails.ucas.ac.cn}
        }
\thanks{$^{4}$ Beihang University, Beijing, China
         % {\tt\small (lihaosheng23,)@mails.ucas.ac.cn}
        }
\thanks{* indicates contributed equally to this work. $\dagger$ indicates project lead. $\ddagger$ indicates corresponding author.}
}

% \begin{document}

\maketitle
\thispagestyle{empty}
\pagestyle{empty}

%%%%%%%%%%%%%%%%%%%%%%%%%%%%%%%%%%%%%%%%%%%%%%%%%%%%%%%%%%%%%%%%%%%%%%%%%%%%%%%%
\begin{abstract}
In real-world scenarios, multi-view cameras are typically employed for fine-grained manipulation tasks. Existing approaches (e.g., ACT~\cite{AcT}) tend to treat multi-view features equally and directly concatenate them for policy learning.
However, it will introduce redundant visual information and bring higher computational costs, leading to ineffective manipulation.
% Fine-grained manipulation tasks typically involve multiple stages, wherein the most advantageous view for different stages varies over time. 
Fine-grained manipulation tasks typically consist of multiple stages, where the best view may vary across different phases.
This paper proposes a plug-and-play Best-Feature-Aware (BFA) fusion strategy for multi-view manipulation tasks, which is adaptable to various policies. Building upon the visual backbone of the policy network, we design a lightweight subnetwork to effectively predict the importance score of each view. 
% ensuring efficient integration.
Based on the predicted importance scores, the reweighted multi-view features are subsequently fused and fed into the end-to-end policy network for seamless integration. Notably, our method demonstrates outstanding performance in fine-grained manipulations. The experimental results show that our approach outperforms multiple baselines by \textbf{22-46\%} success rate on different tasks. Our work provides new insights and inspiration for tackling key challenges in fine-grained manipulations.
\end{abstract}

% \begin{IEEEkeywords}
% Fine-grained Manipulation, Robotics, Dynamic View
% \end{IEEEkeywords}

%%%%%%%%%%%%%%%%%%%%%%%%%%%%%%%%%%%%%%%%%%%%%%%%%%%%%%%%%%%%%%%%%%%%%%%%%%%%%%%%

\begin{figure*}[h]
    \centering
    \includegraphics[width=1.0\textwidth]{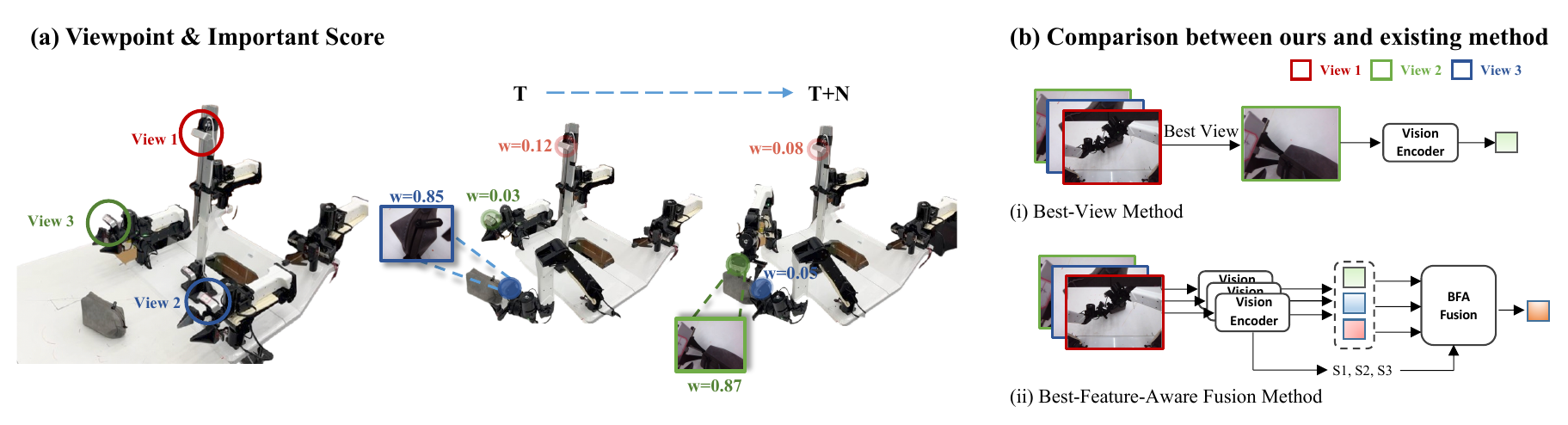}
    \vspace{-0.9cm}
    \caption{(a) Our BFA method evaluates the importance score of each view during manipulation, as indicated by $w$. (b) Unlike the common best-view-based methods, our approach does not simply select the best view. Instead, it performs feature fusion based on the predicted importance scores $S1,S2,S3$ to derive the most suitable observation features for the manipulation task.}
    \label{fig:overview}
    \vspace{-0.5cm}
\end{figure*}

\section{INTRODUCTION}

Imitation learning~\cite{ho2016generativeadversarialimitationlearning, imitation-survey} for robot manipulation~\cite{AcT,rdt, rvt,rvt-2} enables robots to learn and replicate operations from human demonstration data. 
While existing approaches achieve promising results in general manipulation tasks, fine-grained manipulation remains particularly challenging due to its demand for high precision.
Addressing these challenges requires comprehensive scene understanding, which demands multi-view observations to capture both the global context and detailed local interactions.

Notably, the importance of these views varies significantly across different stages of the manipulation process. As shown in the Figure.~\ref{fig:overview} (a), When a robotic arm initially approaches an object, the top view capturing the entire scene becomes crucial, providing essential information about global spatial relationships, scene layout, and target object positioning - while the wrist cameras may not even have the target objects in view. The emphasis shifts markedly during fine-grained manipulation like precise grasping or insertion, where the head or top camera view becomes invaluable by capturing detailed local interactions between the end-effector and target object, enabling precise alignment and depth perception. Therefore, it is important to switch view dynamically during the manipulation process, which can help the model mainly focus on the best camera view at each manipulation stage to better capture crucial spatial and contextual information and enhance operational precision. However, existing methods~\cite{AcT,rdt,ChiDP2023,rvt,rvt-2} typically adopt oversimplified strategies that treat all views as equally important. They either simply concatenate features from different views or directly stack multiple images, neglecting the dynamic significance of each view. This uniform treatment overlooks the evolving importance of different views during manipulation and may bring a significant amount of distracting, unnecessary information, finally leading to reduce manipulation effectiveness and precision.

In this paper, we propose a novel learning-based Best-Feature-Aware (BFA) fusion strategy to address this often overlooked challenge. Our framework dynamically predicts the importance scores of multiple viewpoints by assessing the current interaction state between the robotic arm and objects. Specifically, a lightweight Score Network is introduced to evaluate the significance of each view.
% , enabling real-time view transfer during robot operations. 
Based on the predicted importance scores which can be viewed as signal-to-noise ratios (SNR) of each view, we reweight and fuse the multi-view features, ensuring that the most useful information is effectively integrated to enhance policy performance as shown in the Figure.~\ref{fig:overview} (b). 
% The use of addition allows for a more comprehensive fusion, avoiding the risk of significant information loss. Additionally, the continuous adaptation of feature weights ensures a smoother and more stable integration.
This plug-and-play component enhances the interaction with the environment through adaptive visual perception.

% Based on these importance scores, we then performs weighted fusion of features extracted from different observational views, as illustrated in the Figure.~\ref{fig:BFA-main}

%LHS: 
Furthermore, we designed an automated annotation framework using Vision-Language Models (VLM) that produces the multi-view ground-truth of importance score.
% It can dynamically evaluate the importance of different views in robotics tasks. 
Our system analyzes linguistic and visual proprioception information to categorize the current \textbf{state} of each robotic hand into one of four states:  ``holding",  ``approaching",  ``operating" and ``returning"  . Through carefully designed task-specific rules for combining these states as shown in the Tab.~\ref{tab:hard-rules}, we decompose the entire manipulation process into distinct \textbf{stages}, with each stage focusing different views. We use the annotations from the VLM annotation system to train the above Score Network. 
Our method is evaluated on bimanual manipulation platform ALOHA~\cite{AcT} in real world as shown in Figure.~\ref{fig:overview}(a). The effectiveness of our BFA strategy is further validated on two typical state-of-the-art imitation learning methods, RDT~\cite{rdt} and ACT~\cite{AcT}.
Remarkably, our method demonstrates outstanding performance across various complex fine-grained manipulation tasks, such as ``unzipping bag” and ``opening box”.   It achieves a success rate improvement of \textbf{22-46}\% with existing methods. Moreover, the proposed BFA strategy can reduce the overall computation burden thanks to dynamic view selection. BFA can be regarded as a dynamic network from the perspective of visual perception, improving the effectiveness of policy approaches. We hope the BFA mechanism can provide some new insights for the robotic manipulation community.
% The key contribution of this paper is the plug-and-play BFA framework. The framework significantly improves the success rate in multi-view fine-grained manipulation compared to baseline approaches, Greatly improving the success rate of baseline in fine-grained complex scenarios.
\section{RELATED WORK}

\subsection{View Planning in Manipulation}

View planning in robotics has been widely introduced in \cite{DBLP:journals/cvm/ZengWZL20}, \cite{chen2011active}, which seeks to determine the maximum information gain viewpoint and ensure the sequence of sensors. Among the various domains of view planning, one key area of interest is manipulation , which has been explored through various approaches to optimize task performance.
Arruda et al. \cite{ArrudaDexterous2016} proposed a geometry-based method that prioritizes object visibility and graspability, improving both the quality of reconstruction and the success of grasp. Alternatively, Jun Lv et al. \cite{lv2023samrlsensingawaremodelbasedreinforcement} introduced the differentiates between the manipulation arm and viewpoint arm, magnifying the operation area through viewpoint following to enhance grasping stability. 
Other methods~\cite{DBLP:journals/corr/abs-2409-17435,BreyerNBV2022,WangOTA2024} selected the view with the greatest information gain during operation to address issues such as occlusion.
In recent years, learning-based approaches have been increasingly used to optimize view planning. Some approaches~\cite{cheng2018reinforcement, ShangR2023, chen2020transferable} optimizes viewpoints planning process via reward functions during manipulation.
Additionally, Multi-View Picking~\cite{zaky2020active} applied a self-supervised state representation methods to focus on the target by changing views, enabling the completion of complex manipulation tasks.

\begin{figure*}[htbp]
    \centering
    \vspace{0.3cm}
    \includegraphics[width=1\textwidth]{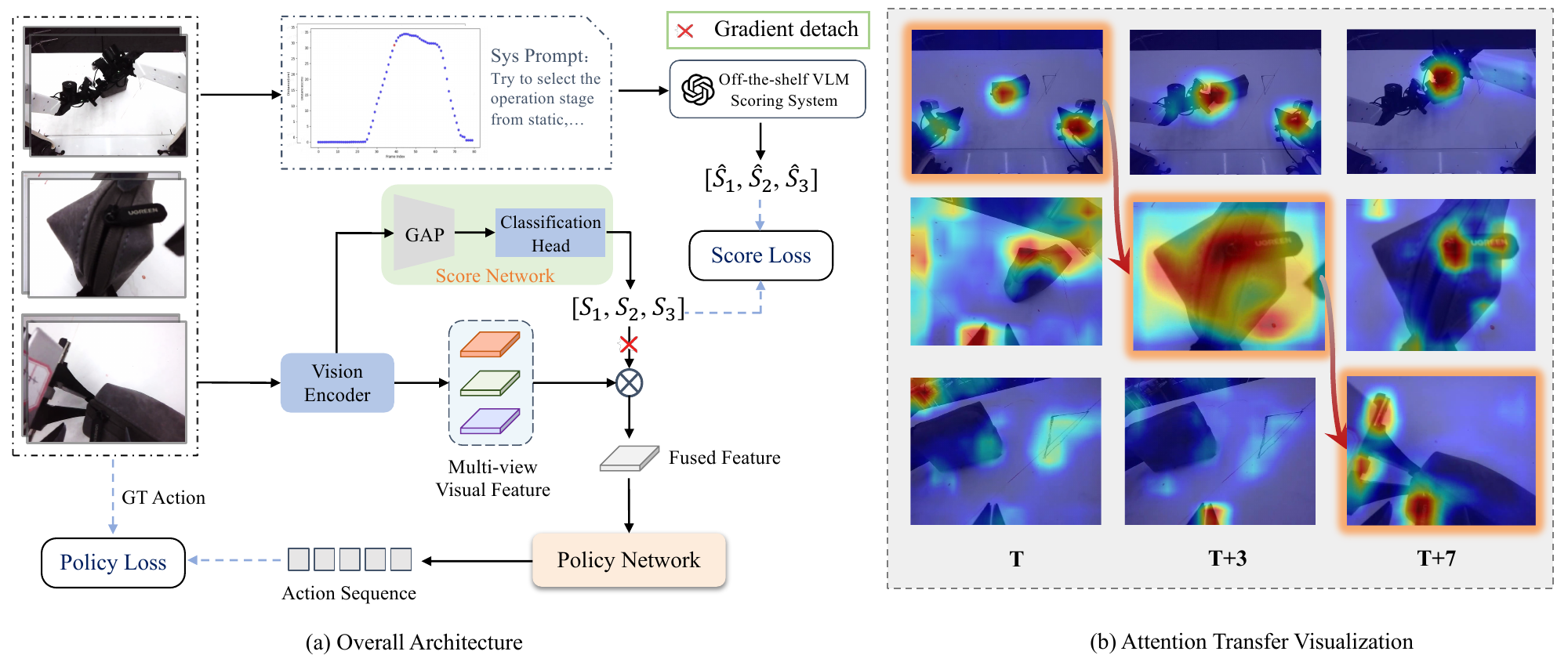}
    \vspace{-0.5cm}
    \caption{The overall pipeline of best feature aware fusion strategy applied in the end-to-end policy network. Top-view and wrist-camera images are first fed into visual backbones to extract per-view features. Those features go into a lightweight scoring network, which assigns an importance weight to each view. The importance scores are further used to reweight and fuse the multi-view features. The fused features are finally served as the input of the policy network to generate the action sequence for real-arm deployment. Moreover, an off-the-shelf VLM annotation system is used to annotate the importance scores of each view offline.} During training, the whole network is jointly optimized by the score loss and the policy loss.
    \label{fig:BFA-main}
    \vspace{-0.3cm}
\end{figure*}

\subsection{Fine-Grained Robotic Manipulation}

Current methods often employ imitation learning strategies to complete fine-grained manipulation tasks. By leveraging expert demonstrations, imitation learning enables the agent to efficiently acquire complex skills.
% \cite{ZhaoALOHA2023, ChiDP2023, Ze2024DP3, NemecVMA2018, KramPass2018, KhorArm2020}. For instance
Some methods~\cite{AcT,rvt,rvt-2}, proposed an imitation learning framework based on transformer architecture \cite{VaswaniAttention2017}, leveraging multi-view information and joint data as demonstration inputs to predict future action sequences. Additionally, Some works~\cite{ChiDP2023,Ze2024DP3,zhao2024aloha} integrate the diffusion process into imitation learning. Moreover, Some works~\cite{bharadhwaj2024roboagent,ma2024hierarchical} have introduced a multi-task approach within these two paradigms, aiming to use a single model for handling multiple tasks.
However, all these methods integrate multi-view information by directly concatenating all visual representations, without considering the unequal information provided by different viewpoints.

\section{METHODOLOGY}

In this section, we will first describe the overall architecture of BFA applied in existing policy networks (see Sec. \ref{sec:overall}). Then we will provide the detailed implementation of BFA, as shown in Sec. \ref{sec:BFA}. After that, the VLM annotation system is presented to generate the ground-truth of importance scores, which is shown in Sec. \ref{VLM Anno}. In Sec.~\ref{analysis}, we provide an in-depth analysis of the mechanism behind the effectiveness of BFA.

\subsection{Overall Architecture}
\label{sec:overall}

Our proposed best-feature-aware (BFA) fusion method is a general strategy which can be viewed as a plug-and-play module used in different end-to-end imitation learning methods. As shown in Fig.~\ref{fig:BFA-main} (a), given multi-view RGB images from top-view and wrist cameras, the vision encoder (e.g., ResNet-18~\cite{HeResnet2016}, SigLIP ~\cite{zhai2023sigclip}) extracts the multi-view visual features respectively. BFA takes the multi-view features as the inputs and uses Score Network (e.g., Multi-Layer Perceptions) to generate the importance score for each view. The generated importance scores are used to reweight and fuse the multi-view features, producing one fused feature which also is the input of subsequent policy network. The policy network (e.g., ACT~\cite{AcT}, RDT~\cite{rdt}) predictsw the action sequences for the deployment of real arm. During training, the Score Network is supervised by the importance score generating from the vision-language model (e.g., GPT-4o~\cite{hurst2024gpt4o}).

\subsection{Best Feature Aware}
\label{sec:BFA}

% add motivation
Previous studies~\cite{AcT,rdt,pi0,robomatrix,octo_2023} typically treat images from different views equally. We observe that each view contributes differently at various stages of the task. Accordingly, we argue that views should not be treated uniformly; instead, the most informative ones should play a dominant role during their most relevant stages.
While primary features typically carry critical information, secondary features often provide complementary cues that further improve performance. Discarding them may limit the model’s performance, so we propose retaining selective secondary features via a Best-Feature-Aware (BFA) fusion  strategy.

Usually, existing imitation learning policies tend to employ the vision backbone (e.g., ResNet-18, SigLIP~\cite{zhai2023sigclip}) to extract the visual features.
% We utilize ResNet-18 or SigLIP as the backbone to extract features from each input view, depending on whether the policy network is ACT-based or RDT-based. 
Let $f_i \in \mathbb{R}$ represent the feature vector extracted from the $i$-th view, where $i \in {1,2, \dots,N}$
and $N$ is the total number of views. The vision backbone processes each view respectively and generates a set of multi-view features:
$$\mathcal{F} = \{ f_1, f_2, \dots, f_N \}.$$

To achieve prioritized viewpoint selection, we reuse the multi-view visual features and design a plug-and-play lightweight network $F_{cls}$ to predict the importance scores for each view.
% To achieve prioritized viewpoint selection, we designed a plug-and-play network $F_{cls}$. 
For both ACT~\cite{AcT} and RDT~\cite{rdt}, the multi-view features $\{f_1, \ldots, f_N\}$ are compressed to low-dimensional representations via global average pooling ($GAP$). 
Then we fed the extracted feature to the Class Head $F_{cls}$ to predict the importance score, \( s_1, \ldots, s_N \) for all views.
\begin{equation}
    s_i=F_{cls}(GAP(f_i))
    \label{equal:forward}
\end{equation}
In our practice, we employ a three-layer linear network as Class Head to predict the importance scores.
To better integrate multi-view feature information, not only the visual feature of the most important view is used as the observation feature of policy network. 
% Instead, we retain the weights for the features. This allows the model to dynamically focus on key views from all views in each frame, producing temporal attention transfer. 
Instead, we utilize the predicted importance scores to reweight the features of corresponding views and then perform the element-wise addition operation to fuse the reweighted multi-view features.
\begin{equation}
   \hat{f} = \sum_{i=1}^N f_{i} \times s_i
    \label{equal:fusion}
\end{equation}
Here, $\hat{f}$ is the fused visual representation for fine-grained manipulation. 
The fused feature $\hat{f}$ is then fed into subsequent policy network for action sequence prediction.
This element-wise addition allows for a more comprehensive fusion, avoiding the risk of the total information loss of unimportant views. Additionally, the continuous adaptation of feature weights ensures a smoother and more stable integration.

The policy loss $L_{p}$ is used to optimize the parameters of policy networks as well as the visual backbone.
Correspondingly, we add the score loss $L_{s}$ as a auxiliary task.
\begin{equation}
   {L}_{\text{s}}=BCE(s,\hat{s})
    \label{equal:bce}
\end{equation}
where $BCE$ is the binary cross-entropy loss and $\hat{s}$ is the ground-truth of the importance score annotated by the VLM, which is described in Sec. \ref{VLM Anno}.
Therefore, the overall loss function can be formulated as:
\begin{equation}
    L =\lambda_1 L_{\text{s}} + \lambda_2 L_{p}
    \label{equal:total-loss}
\end{equation}
where $\lambda_{1},\lambda_{2}$ represents the weights for two loss functions. During training, the gradients of policy loss are propagated through all components except Score Network as shown in the Figure. \ref{fig:BFA-main}, while the score loss gradients are simultaneously propagated to the vision encoder.

% \begin{figure*}[h]
%     % \centering
%     \includegraphics[width=0.5\textwidth]{figures/Baseline-BFA-gradcam.pdf}
%     \vspace{-0.8cm}
%     \caption{\textbf{}}
%     \label{fig:baseline vs BFA}
% \end{figure*}

\subsection{VLM Scoring System}
\label{VLM Anno}

\begin{figure}[h]
    \centering
    \includegraphics[width=0.5\textwidth]{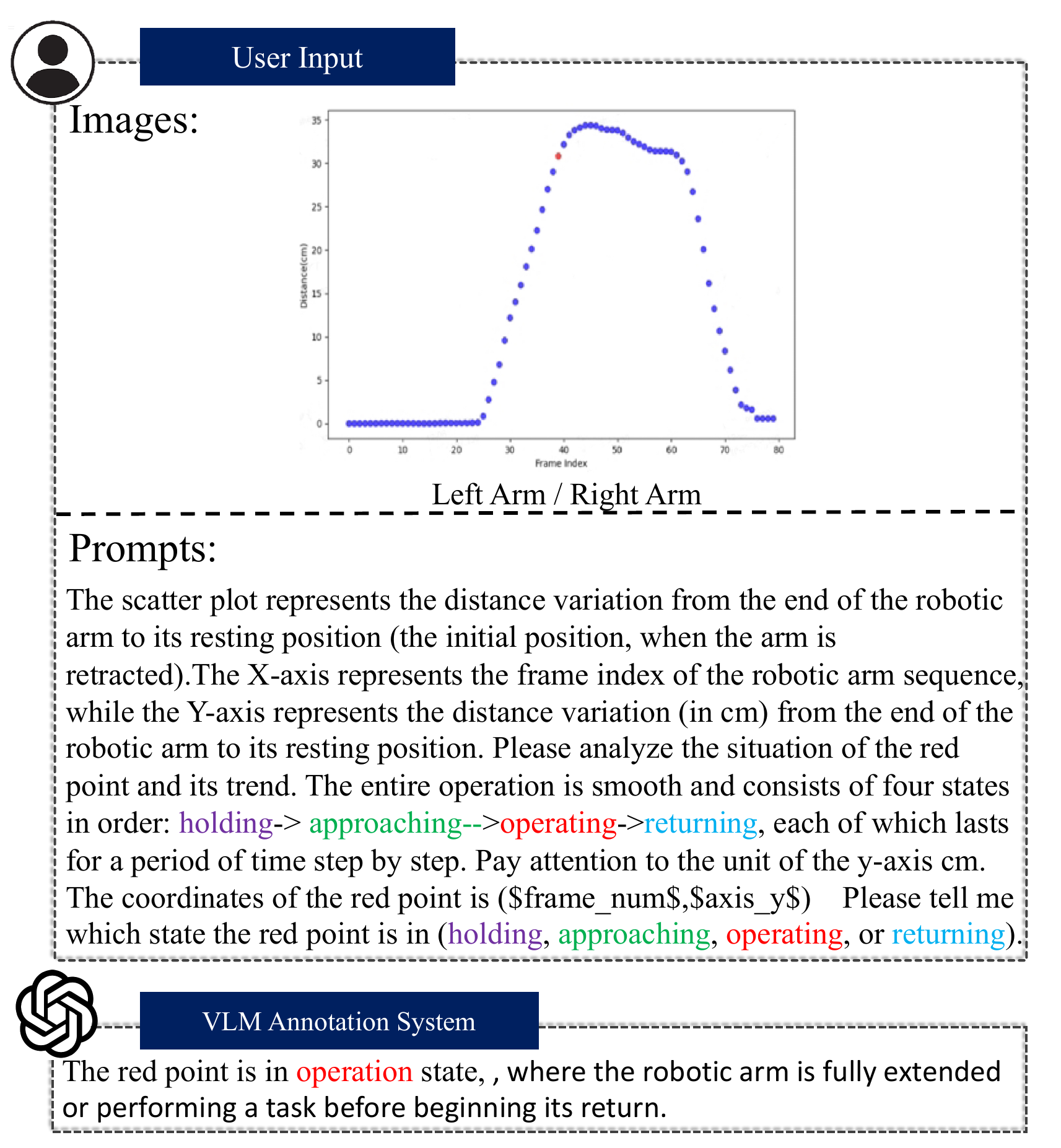}
    \vspace{-0.65cm}
    \caption{System Prompt of VLM Scoring System. Given the plotted images and system prompt, the VLM system outputs current state of the robotic arm.}
    \label{fig:BFA-prompt}
    \vspace{-0.3cm}
\end{figure}

To generate the ground-truth of importance score, we develop a VLM scoring system.
% In our scenario, we have $n=3$ viewpoints, consisting of two wrist-mounted views and one front view. 
To achieve human-level annotation quality in our system, we combine the rule-based methods with a Vision-Language Model (VLM) \cite{gpt-4o}. Using proprioceptive information from the robotic arms, we divide the entire episode into different stages
%\blue{select} out transition phases from the entire sequence. 
For each frame, we generate two scatter plots that represent the change in distance from the grippers of the left and right robotic arms to their respective resting positions over time. These scatter plots are then input into the VLM \cite{gpt-4o}, which is queried to determine the current state of each robotic arm (e.g. ``holding", 
 ``approaching",  ``operating" and ``returning"  ), as shown in Fig.~\ref{fig:BFA-prompt}.

\begin{figure*}[h]

    \centering
    \includegraphics[width=1\textwidth]{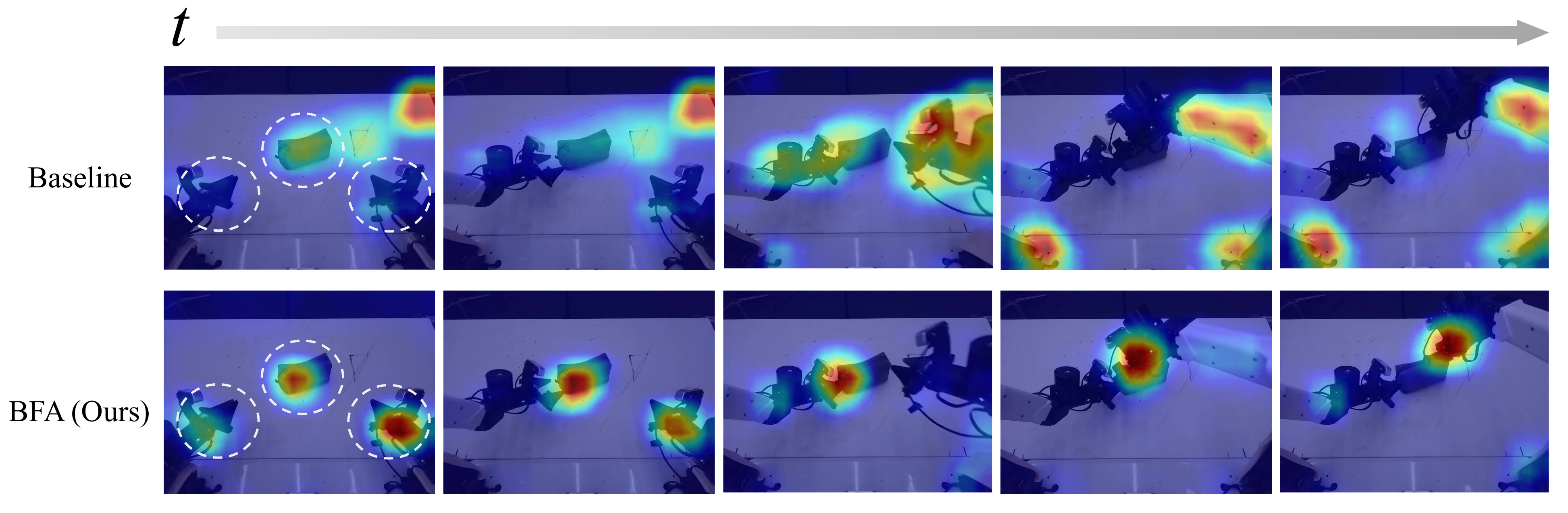}
    \vspace{-0.65cm}
    \caption{Grad-Cam Heatmap Comparision between baseline and BFA. Our BFA affords more attention on end-effector and target object edge across the wall-time, while baseline distracts in the manipulation process.}
    \label{fig:BFA-gradcam-comparision}
\vspace{-0.3cm}
\end{figure*}

To identify the state of both the left and right robotic arms, we combine the phases of both arms and apply hard rules which is shown in the Tab.~\ref{tab:hard-rules} to calculate the view score of our system for the different viewpoints at that moment. For instance, if the left arm is in a ``holding" state while the right arm is in an ``operating" state, more attention should be given to the right-hand view. In this scenario, the importance score would be [0, 0, 1], allowing the system to focus mainly on the right-hand view. Furthermore, to optimize resource usage, we assume that during transitions — when the robotic arms move from the resting position towards an object or return to the resting position — the focus should shift to the top camera view. In this case, the importance score would be [0, 1, 0].

%\blue{\textbf{Comparision between Rule and VLMs.}} \red{\textbf{...}}

\vspace{-0.3cm}

\vspace{-0.3cm}
\renewcommand{\arraystretch}{1.3} % 调整行高，默认值是 1
\setlength{\tabcolsep}{4pt}
\begin{table}[h!]
\centering
\caption{State and Stage Rules of Some Tasks}

\begin{tabular}{|c|c|c|c|}
\hline
\rowcolor{blue!40} \textbf{Task Name} & \textbf{Left Arm State} & \textbf{Right Arm State} & \textbf{Important Score} \\
\hline
\multirow{5}{*}{Unzip bag} & Approaching & Holding & Top [0,1,0]\\
\cline{2-4}
                            & Operating & Holding & Left [1,0,0]\\
\cline{2-4}
                            & Operating & Approaching & Top [0,1,0]\\
\cline{2-4}
                            & Operating & Operating & Right [0,0,1]\\
\cline{2-4}
                            & Returning & Returning & Top [0,1,0]\\
\hline

\multirow{3}{*}{Fold towel} & Approaching & Approaching & Top [0,1,0]\\
\cline{2-4}
                            & Operating & Operating & Left\&Right [1,0,1]\\
\cline{2-4}
                            & Returning & Returning & Top [0,1,0]\\               
\hline
\multirow{3}{*}{Open box} & Approaching& / & Top [0,1,0]\\
\cline{2-4}
                            & Operating & / & Left [1,0,0]\\
\cline{2-4}
                            & Returning & / & Top [0,1,0]\\               
\hline
\end{tabular}
\label{tab:hard-rules}
\end{table}
\renewcommand{\arraystretch}{1.1}
Additionally, we implement a frame-skipping annotation strategy in which we annotate every five frames. If the annotations before and after the 5-frame window are consistent, we apply the same annotation to all frames within that window. However, if there is an inconsistency, we perform frame-by-frame annotation for the 5 frames. Combining these two optimization methods significantly reduces the computation time required for annotation.

\subsection{Mechanism Analysis}
\label{analysis}

To explore the mechanism of Best-Feature-Aware (BFA), we further provide an in-depth mechanism analysis on the strategy to explain why it works for fine-grained manipulation tasks. Considering that the introduced score network is essentially a classification network, we conduct the attention visualization on three views using Grad-CAM~\cite{selvaraju2017gradcam}. The core of CAM is that its attention region tend to focus on the target object with the highest classification score.

As shown in Fig.~\ref{fig:BFA-main} (b), the visualization of three views shows that the attention of score network is shifted along the temporal axis. For the $T$ timestep when the gripper accesses the manipulated objects, the attention focuses mainly on the gripper and objects from the top view. For the $T+3$ timestep when the left arm operates on the bag, the attention shift from the top view to the left wrist view. Finally, the attention focuses on the gripper of right arm when right arm unzips the bag. The predicted importance score also indicates the best feature transfer (red line). Furthermore, Fig.~\ref{fig:BFA-gradcam-comparision} shows that our method exhibits stronger responses in Regions-of-Interest (RoI)~\cite{ren2016faster}, such as the gripper and target objects, whereas the baseline method displays more dispersed attention.
The BFA strategy achieves best-view selection and transfer by predicting the importance score. 

\vspace{-0.3cm}
\setlength{\tabcolsep}{8pt}
\begin{table*}[ht]
% \caption{Quantitative comparison on different fine-grained manipulation tasks.} 
\caption{Success rate comparison of different methods across multiple fine-grained manipulation tasks}
\centering 
\begin{tabular}{c|c|ccc|cc|cc|cc|cc}
% \begin{tabular}{l|ccc|cc|ccc}
  \specialrule{1pt}{0pt}{1pt}
  \toprule 
  \multirow{2}{*}{Method} & \multirow{2}{*}{Average Suc.} & \multicolumn{3}{c|}{Unzip bag} & \multicolumn{2}{c|}{Play Chess} & \multicolumn{2}{c|}{Open box} & \multicolumn{2}{c|}{Close box} & \multicolumn{2}{c}{Fold towel} \\
  % \multirow{2}{*}{Method} & \multicolumn{3}{c|}{unzip(slid) bag} & \multicolumn{2}{c|}{insert coin} & \multicolumn{3}{c}{place chess} \\
  % \cmidrule(r){3-10}
  \cline{3-13}
   &  & Grab & Pinch & Slide & Pick & Position & Align & Lift& Align & Flip & Grab & Fold\\ 
 % \cmidrule(r){1-1}
 % \cmidrule(r){2-3}
 % \cmidrule(r){4-5}
 % \cmidrule(){6-15}
\midrule

 ACT & 32\% & \textbf{100\%} & 60\% & 30\% & 40\% & 20\% & 50\% & 30\% & 20\% & 30\% & 50\% & 50\%\\ 
 
 \rowcolor[gray]{.9}ACT-BFA (Ours) & \textbf{78\%} & \textbf{100\%} & \textbf{90\%} & \textbf{70\%} & \textbf{100\%} & \textbf{80\%} & \textbf{100\%} & \textbf{90\%} & \textbf{70\%} & \textbf{80\%}&\textbf{80\%} & \textbf{70\%}\\
\hline
  RDT & 20\% & 50\% & 30\% & 20\% & 10\% & 0\% & 50\% & 40\% & 20\% & 20\% & 30\% & 20\% \\ 
 
 \rowcolor[gray]{.9}RDT-BFA (Ours) & \textbf{42\%} & \textbf{80\%} & \textbf{80\%} & \textbf{30\%} & \textbf{50\%} & \textbf{0\%} & \textbf{70\%} & \textbf{70\%} & \textbf{70\%} & \textbf{40\%} & \textbf{80\%} & \textbf{70\%} \\
 % \rowcolor[gray]{.9} w/ Robotics Pretrain &  100\% & 100\% & 100\% & 100\% & 100\% & 100\%\\
  \bottomrule
  
  % \specialrule{1pt}{1pt}{0pt}
 \end{tabular}
 \vspace{-0.5cm}
 \label{tab:pretrain} %\vspace{-0.5em}
\end{table*}

\section{EXPERIMENTS}

\subsection{Robot Setup and Tasks details}

%We conducted experiments to evaluate its performance in fine-grained manipulation tasks as shown in the Tab.~\ref{tab:tasks}. 

\begin{figure*}[h]
    \centering
    \includegraphics[width=\textwidth ]{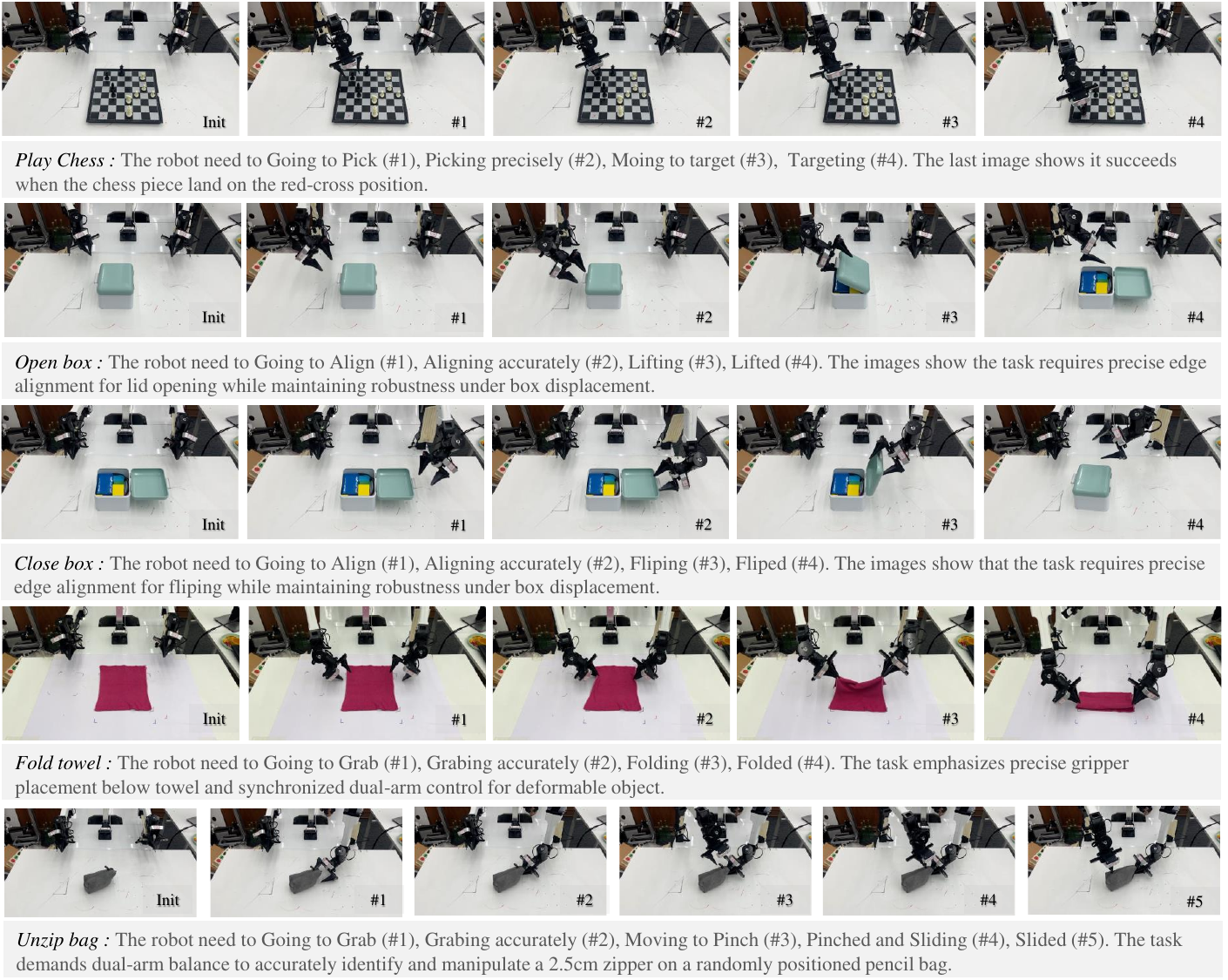}
    \vspace{-0.8cm}
    \caption{\textbf{Visualization of robotic manipulation sequences for five tasks}: playing chess, opening a box, closing a box, folding a towel and unzipping a bag. Each sequence highlights the robot's precise control in alignment, grasping, flipping, and dual-arm coordination, demonstrating the effectiveness of our method in fine-grained manipulation tasks.}
    \label{fig:demo}
    \vspace{-0.3cm}
\end{figure*}

In real world, we build five dexterous manipulation tasks with \textit{ALOHA} which is an open-source Cobot Magic platform. As shown in Fig.~\ref{fig:demo} , among the 5 challenging tasks, two require dual-arm coordination (\textit{\textbf{Fold Towel}}, \textit{\textbf{slide Bag}}) while three involve single-arm manipulation (\textit{\textbf{Play Chess}}, \textit{\textbf{Open Box}}, \textit{\textbf{Close Box}}).
This platform includes four AgileX robotic arms and four Orbbec DaBai RGB cameras mounted on the Tracer chassis. The camera mounted on the top reveals the global view, while each arm is equipped with a wrist camera, as described in the Fig. \ref{fig:overview}(a). All cameras capture images at 30Hz frequency. At each timestep, the robotic system captured frames from the cameras, each delivering 640×480 pixels RGB images.

% \setlength{\tabcolsep}{1.8pt}
% % \vspace{-0.5cm}
% \begin{table}[h!]
% \caption{Task Details}
% \centering
% \begin{tabular}{c|c c|c|c}
% \hline
% \textbf{Task} & \textbf{Dual Arm} & \textbf{Random Place} & \textbf{ACT Episodes} & \textbf{RDT Episodes} \\
% Unzip bag & \checkmark & \checkmark & 50 & 300 \\
% Play Chess &  &  & 50 & 50 \\
% Open Box &  & \checkmark & 50 & 500 \\
% Close Box &  & \checkmark & 50 & 500 \\
% Fold towel & \checkmark & \checkmark & 50 & 500 \\
% \hline
% \end{tabular}
% \label{tab:tasks}
% \end{table}
% \setlength{\tabcolsep}{4pt}
% \vspace{0.3cm}

\subsection{Implementation Details} 

For all five real-world tasks, we collected demonstrations with each episode requires 300 to 650 timesteps to perform a complex task, given the control frequency of 25 Hz. We record 50 to 500 episodes for each task. 
%as shown in the Tab.~\ref{tab:tasks}. 
We recorded the average success rate on the fine-grained manipulation tasks. For each task, we conducted ten trials using original ACT policy and RDT policy, respectively. 

% All experiments were conducted using NVIDIA hardware. 
Both networks were trained from scratch with random initialization, without pre-training or fine-tuning. The ACT model integrated with BFA method contains approximately 106M parameters and is trained independently for each task. Training completes in approximately 2 hours on a single NVIDIA RTX 4090 GPU, achieving an inference time of 0.015 seconds on an RTX 4060 GPU (8GB). The RDT model with BFA method, comprising around 170M parameters, is trained from scratch per task on NVIDIA A100 GPUs. The training process takes 3 hours using 8 A100 GPUs (80G) in parallel, or 1 hour on a single A100 GPU, with an inference time of approximately 0.73 seconds on an RTX 4060 GPU.

%\blue{\textbf{Action Chunking Size.}} 
We implement ACT with the same action chunking size as to ACT-BFA, which is 24. RDT and RDT-BFA's chunk size are 64.

\subsection{Experiment Results}

As shown in Tab.~\ref{tab:pretrain}, We integrated our modules into two state-of-the-art imitation learning methods ACT~\cite{AcT} and RDT~\cite{rdt} and conducted comparative experiments.
ACT-BFA and RDT-BFA consistently outperform the baseline models in success rates at every stage across the five tasks. Overall, both models show significantly higher success rates, with ACT-BFA achieving a 46\% improvement and RDT-BFA showing a 22\% increase compared to their respective baseline models.

Overall, the low success rate of the baseline models is primarily due to the inability to locate positions accurately and precisely control the gripping. This is because, when performing fine-grained manipulation, multiple views introduce excessive redundant information, which makes it challenging for baseline methods~\cite{AcT, rdt} to accurately and generally learn the mapping pattern from visual features to future actions. To validate this reasoning, we analyzed the failure cases of ACT~\cite{AcT} and RDT~\cite{rdt}. Both methods exhibit similar failure patterns when generalization requirements are introduced, such as varying box positions and random grasp points on deformable objects. In these cases, both policies show significant deviations in estimating critical grasp points, which ultimately hinder progression to subsequent stages.

Moreover, the two baselines differ in the types of failures. ACT~\cite{AcT} performs well in basic tasks but fails during fine manipulation due to insufficient trajectory refinement. For example, in the \textit{Unzip bag} task, after securing the bag, the right gripper often misaligns with the zipper, leading to failure. On the other hand, RDT~\cite{rdt} experiences minor yet persistent trajectory errors across all stages, causing issues like bimanual coordination failures. For instance, in the \textit{Fold towel} task, one gripper may lift the towel’s corner while the other fails to grasp it properly, yet continues the folding motion. All these observed failure behavior aligns with our hypothesis regarding the limitation of baseline methods.

In contrast to these issues, our method effectively resolves the challenges through view selection and feature fusion, and can accurately identify the gripper point even in generalized settings, while autonomously correcting errors during the manipulation process, as demonstrated in Tab.~\ref{tab:pretrain}.

\subsection{Ablation Study}
In this section, we present an ablation study on ACT-BFA to investigate the impact of various design choices for fusing and the number of views during the manipulation. We focus on the fine-grained manipulation task of ``Unzip Bag" and ``Open box'' as the experimental setting for our ablation analysis.

To reveal the importance of continuously adjusting the fusion weights during manipulation process, we employed four different fusion strategies which named ``Mean", ``Reweight Concat", ``Best Feature" and  ``w/o Score Loss" as shown in the Tab.~\ref{tab:ablation_selection}. 
For ``Mean", We simply average the visual features from multiple perspectives as defined by $\hat{f} = \frac{\sum_{i}^{N} f_i}{N} $.

For ``Reweight Concat", we multiply the scores obtained from the scoring network with the visual features, then directly concatenate them and pass the result to the policy network as defined by $\hat{f} =[ f_{1} \times s_1,...,f_{N} \times s_N ]$. 
For ``Best Feature", we use the scores from the scoring network to select the visual feature with the highest score, which is then passed to the policy network following $\hat{f} =f_{arg max(s_1,...,s_N)}$. For ``w/o Score Loss," we rely solely on policy loss for supervision, enabling the model to learn an importance score without the need for human-provided ground truth.

These four methods represent other commonly used fusion approaches. The results is presented in Tab.~\ref{tab:ablation_selection}. Our fusion strategy demonstrated significantly higher success rates on the unzip task compared to other methods, which proves the effectiveness of our fusion strategy.The success rate of the `Mean' fusion method and the baseline remains consistent in the end. 
\vspace{-0.5cm}
\setlength{\tabcolsep}{8pt}
\begin{table}[h]
    \caption{{Ablation study on feature fusion.}}
    \centering
    \begin{tabular}{@{}c|ccc|cc@{}}
    \toprule
        \multirow{2}{*}{Fusion Method} & \multicolumn{3}{c|}{Unzip Bag} & \multicolumn{2}{c}{Open Box} \\[2pt]
        \cline{2-6} 
         & Grab & Pinch & Slide & Align & Lift  \\
        \midrule
        Mean & 90\% & 60\% & 30\% & 50\% & 30\% \\
        Best Feature& \textbf{100\%} & 80\% & 40\% & 80\% & 60\% \\
        Reweight Concat & 80\% & 60\% & 40\% &  50\% & 40\% \\
        w/o Score Loss & 70\% & 40\% & 20\% & 80\% & 70\% \\
        \midrule
        Ours & \textbf{100\%} & \textbf{90\%} & \textbf{70\%} & \textbf{100\%}  & \textbf{90\%} \\
    \bottomrule
    \end{tabular}
    \vspace{-0.3cm}
    \label{tab:ablation_selection}
\end{table}

Moreover, it is worth noting that the latter three methods which utilize the important score ultimately outperform the baseline, indicating that the score network effectively guides the model to better learn from the human demonstrations. 
Since the `Best Feature' approach directly selects the view with the highest importance score, it results in a significant loss of information, which negatively impacts both generalization and policy performance. Moreover, `Best Feature'’s discrete feature weight selection results in large fluctuations in unseen scenarios, whereas our method’s continuous weight adaptation ensures smoother, more stable decision-making.
Moreover, `Reweight Concat' approach fails to effectively fuse features as our method does. It merely applies weighted multiplication without true integration, offering only minimal improvement over the baseline. Our approach, however, leverages prior knowledge by pre-setting signal-to-noise ratios for each view, enabling more efficient weighting and better utilization of available information. In comparison, Reweight Concat shifts the fusion responsibility to the policy itself, which can be less effective.
Finally, the ``w/o score loss" method allows the model to learn the importance score independently, and it is much less effective than our approach. The important score remains fixed across manipulation stages, such as $[0.4, 0.2, 0.4]$ for unzip bag and $[0.4, 0.6]$ for open box. This approach is essentially a simple weighted feature fusion with fixed weights, lacking our method's ability to adapt to different manipulation stages. 
%\blue{Why w/o Score Loss Performs bad? Whether scores too jittery?} 
Its score does not fluctuate violently but only varies within a narrow range; the backpropagated gradient of the action loss is not strong enough to enable it to fuse useful features effectively.

\begin{table}[h]
    \caption{{Ablation study on view selection.}}
    \centering
    \begin{tabular}{c|ccc|cc}
    \toprule
        \multirow{2}{*}{Viewpoint} & \multicolumn{3}{c|}{Unzip Bag} & \multicolumn{2}{c}{Open Box} \\[2pt]
        \cline{2-6} 
           & Grab & Pinch & Slide & Align & Lift  \\
        \midrule
        Top-view& 50\% & 20\% & 10\% & 30\% & 10\% \\
        Left-wrist & 0\% & 0\% & 0\% &  - & - \\
        Right-wrist & 30\% & 20\% & 10\% & 30\% & 10\% \\
        Baseline (3-views) & \textbf{100\%} & 60\% & 30\% &  50\% & 30\% \\
        \midrule
        Ours (3-views) & \textbf{100\%} & \textbf{90\%} & \textbf{70\%} & \textbf{100\%} & \textbf{90\%} \\
    \bottomrule
    \end{tabular}
    \vspace{-0.3cm}
    \label{tab:view_selection}
\end{table}
We then conducted an ablation study on the baseline's view selection as shown in the Tab.~\ref{tab:view_selection}. The results indicate that using only one camera performs worse than using all three views. This suggests that simply reducing the number of views is not a viable strategy—for instance, relying solely on the top view results in limited accuracy during manipulation, and using only the wrist camera fails to achieve precise positioning at close range. Moreover, our method significantly outperforms the configuration that uses all three views, thereby demonstrating its effectiveness.

\vspace{-0.3cm}
\setlength{\tabcolsep}{1.0pt}
% \vspace{-0.5cm}
\begin{table}[h!]
\caption{The Comparison of Different Annotation Method}
\centering
\begin{tabular}{c|c |c|c|c|c|c}
\toprule
\textbf{Method} & \textbf{Unzip bag} &\textbf{Close box} & \textbf{Play Chess} & \textbf{Open Box} & \textbf{Fold towel} & \textbf{Avg} \\
\toprule
Human & 1.00 & 1.00 & 1.00 & 1.00 & 1.00 & 1.00 \\

Rule-based & 0.92 & 0.90 & 0.84 & 0.92 & 0.88 & 0.89\\
\rowcolor[gray]{.9} VLM-based & 0.99 & 0.98 & 0.97 & 0.99 & 0.98 & 0.98 \\
  \bottomrule
\end{tabular}
\vspace{-0.3cm}
\label{tab:anno}
\end{table}
\setlength{\tabcolsep}{4pt}

To validate our annotation method, we compared different annotation approaches against manual human annotations. Due to cost constraints, we evaluated 10 episodes per task. We developed a rule-based method using gradients of distance from both left and right grippers. At the same time, human annotators only labeled dual-arm manipulation states, and both the manual annotations and rule-based method adhered to the same hard mapping rules outlined in Table~\ref{tab:hard-rules}. Using manual annotations as the ground truth, our comparison revealed that the VLM-based approach achieves better alignment with human annotations than the rule-based method, as demonstrated in Tab.~\ref{tab:anno}.

\vspace{-0.3cm}
\begin{table}[h]
  \centering
  \caption{ \textbf{Comparison of success rate and computational cost.}
  }
  \label{tab:speed}
  \begin{tabular}{c|c|c|c}
    \toprule
% num\_channels & mAP & NDS & FLOPs (G) & \# params. (M) \\
Method & Suc.(\%) & FLOPs (G) & \# params. (M) \\
\hline
ACT & 32 & 16.34 & \textbf{106.22} \\ 
ACT-BFA (Ours) & \textbf{78} & \textbf{12.96} & 106.90 \\ 
RDT & 20 & 4356.99 & 166.23  \\
RDT-BFA (Ours) & \textbf{42} & \textbf{3805.66} & \textbf{162.50} \\
% \hline
\bottomrule
  \end{tabular}\vspace{-3mm}
\end{table}

To further demonstrate the effectiveness of our BFA strategy, we report the overall performance, the computational cost and the parameters (see Tab.~\ref{tab:speed}). It shows our BFA only adds marginal parameters, while greatly reducing the computational cost and improving the success rate. The BFA reduces the information redundancy of multi-view images, providing the informative visual features for  policy networks.

\section{CONCLUSIONS}

In this paper, we propose the Best-Feature-Aware fusion strategy for multi-view fine-grained manipulation. Such strategy achieves the  dynamic view fusion during manipulation. It greatly reduces the visual information redundancy and computational costs while significantly improving the success rate of complex fine-grained manipulation tasks. To implement the strategy, we further introduce the VLM-based scoring system to generate the multi-view ground-truth of importance score, as the supervision of the introduced light-weight scoring network. The proposed BFA strategy provides 22\%-46\% improvements on fine-grained manipulation tasks. In the future, we will extend our method to VLA (Vision-Language-Action) based approaches \cite{pi0,openvla,RT-X}. Since most tokens of VLA in these method are image tokens, we anticipate our method's effectiveness and its potential to significantly reduce large computational resources.

% \addtolength{\textheight}{-12cm}   % This command serves to balance the column lengths
                                  % on the last page of the document manually. It shortens
                                  % the textheight of the last page by a suitable amount.
                                  % This command does not take effect until the next page
                                  % so it should come on the page before the last. Make
                                  % sure that you do not shorten the textheight too much.

%%%%%%%%%%%%%%%%%%%%%%%%%%%%%%%%%%%%%%%%%%%%%%%%%%%%%%%%%%%%%%%%%%%%%%%%%%%%%%%%

%%%%%%%%%%%%%%%%%%%%%%%%%%%%%%%%%%%%%%%%%%%%%%%%%%%%%%%%%%%%%%%%%%%%%%%%%%%%%%%%

%%%%%%%%%%%%%%%%%%%%%%%%%%%%%%%%%%%%%%%%%%%%%%%%%%%%%%%%%%%%%%%%%%%%%%%%%%%%%%%%
% \section*{APPENDIX}

%%%%%%%%%%%%%%%%%%%%%%%%%%%%%%%%%%%%%%%%%%%%%%%%%%%%%%%%%%%%%%%%%%%%%%%%%%%%%%%%

% \printbibliography

% \bibliographystyle{ieeetr}
% \bibliography{ieeeconf/IEEEexample.bib}

\bibliographystyle{IEEEtran}
\bibliography{IEEEexample.bib}

\end{document}